\documentclass[runningheads]{llncs}
\usepackage{graphicx}
\usepackage{url}
\usepackage{todonotes}
\usepackage{cleveref}
\usepackage{booktabs}
\begin{document}
\title{A First Experiment on Including Text Literals in KGloVe}


%
%
\author{Michael Cochez\inst{1,2,4} 
	\and
Martina Garofalo\inst{1,3} \and J\'{e}r\^{o}me Len\ss{}en\inst{2} \and Maria Angela Pellegrino\inst{1,3}}
\authorrunning{M. Cochez et al.}
%
\institute{Fraunhofer FIT, Sankt Augustin, Germany
\email{michael.cochez@fit.fraunhofer.de}
\and Information Systems and Databases, RWTH Aachen University, Germany
\email{jerome.lenssen@rwth-aachen.de}
\and Department of Computer Science,	University of Salerno, Italy
\email{\{margar1994,mariaangelapellegrino94\}@gmail.com}
\and Faculty of Information Technology, University of Jyvaskyla, Finland}

%
\maketitle              
\begin{abstract}
Graph embedding models produce embedding vectors for entities and relations in Knowledge Graphs, often without taking literal properties into account. 
We show an initial idea based on the combination of global graph structure with additional information provided by textual information in properties. 
Our initial experiment shows that this approach might be useful, but does not clearly outperform earlier approaches when evaluated on machine learning tasks. 

\end{abstract}
\section{Introduction}
Knowledge Graphs provide ways to organize, manage, and retrieve structured data and give the opportunity to AI systems to perform reasoning. 
A thriving part of artificial intelligence research is focused on the creation of models make predictions based on provided historical data.
Using knowledge graphs as an input to these systems seems a good idea, but exactly how this should be accomplished seems still an active research question.
In this paper, we continue on the branch of work which attempts to enable the use of knowledge graph for machine learning and knowledge discovery by embedding parts of the graph in vector spaces (i.e., graph embedding techniques).
Using these techniques, information from the KG is converted into vectors in a continuous vector space model, with the aim that this embedding would enable the use of existing machine learning models, which typically expect a vector input.

Traditional graph embedding models, such as TransE and TransR, and many newer models produce embedding vectors for each entity and relation in the graph.
However, they do only take the topology of the graph and the labels on the edges into account.
It seems that also including other information, in particular information contained in properties of the nodes in the graph would be a reasonable thing to do.
Especially graphs typically used in the semantic web like, for example DBpedia which we will use in our experiments below, the amount of information available in literals is vast.
Missing this information while embedding seems a disadvantage indeed, but it is unclear how this information should be included into the current models.
Initial investigations on enriching Knowledge Graph embeddings by incorporating literals can be found in several recent related works.
First, LiteralE~\cite{LiteralE} learns embeddings for nodes, edges and literals in the knowledge graph and combines nodes and their corresponding literals with an arbitrary function to produce literal-enriched node embeddings. 
In the work by Toutanova et al.~\cite{TextJointEmbedding}, relations between entities are extracted from text by using predefined patterns. 
The generated triples from text are then jointly trained with the triples in the graph. 
CANE~\cite{CANE} concatenates text embeddings obtained from a CNN with embeddings of entities associated to those literals. 
Furthermore CANE learns for each adjacent entity a different embedding vector for each entity, introducing context awareness. 
A similar approach to the method proposed in this article is used by Type2Vec~\cite{type2vec}, where text from literals is collected and a named entity recognition is performed on the text, removing all non-matching words. 
Then the entities are replaced by their minimal type and word2vec is applied to the text. 
It was observed that word2vec captures the similarity between different types.

The method we present in this paper is based on KGloVe~\cite{KGlove}.
KGloVe works by first creating a co-occurrence matrix from the graph by performing Personalized PageRank (PPR) on the (weighted) graph.
Then, the same optimization as proposed in the GloVe approach~\cite{shortglove} is applied.
In this paper we present a preliminary experiment on our investigation with a new model to incorporate the information contained in literals in a Knowledge Graph into the embedding. 
We perform an experiment, which shows that there might be an advantage from incorporating literal information into latent feature models, but the gain is rather small.

\section{Our Method}
\begin{figure*}
\centering
\includegraphics[width=1\linewidth]{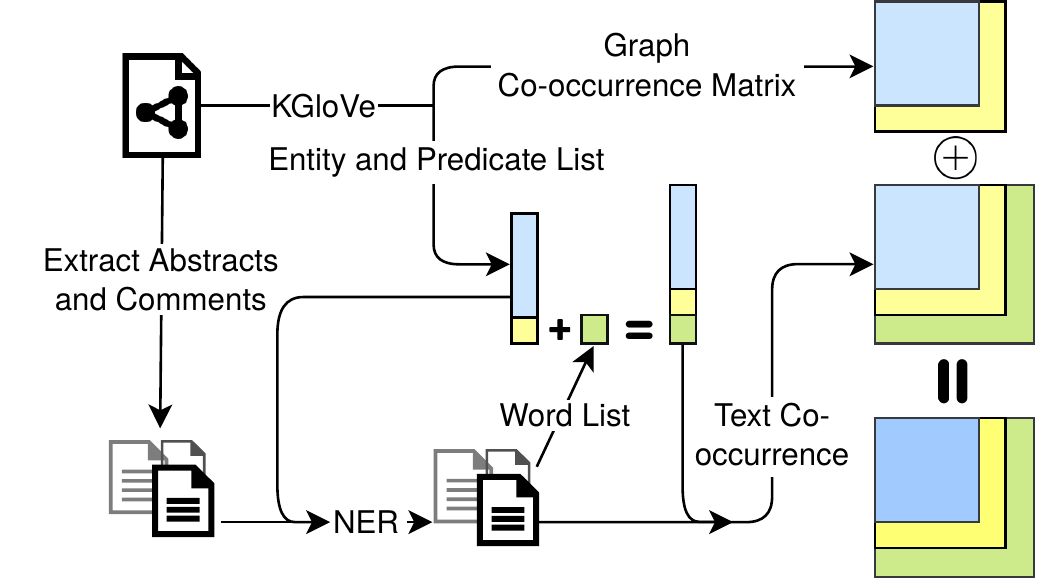}
\caption{Co-occurrence matrix creation workflow}
\label{fig:workflow}
\end{figure*}

Our method is based on KGloVe and extends it to also incorporate the information included in the labels.
Specifically, we use DBpedia abstracts as the technique seems more reasonable for long descriptive texts, rather than short labels like the entity name.
An overview of the approach is presented in \cref{fig:workflow}.

The core idea is to extract information from the abstract by first performing an entity recognition step.
Then, the words representing the entity are replaced by the entity itself and the words (and potentially other entities) surrounding it are also included into the context of the entity.

In detail, starting from the RDF graph, two co-occurrence matrices are created independently and merged at the end.
The first one is obtained as in the KGlove technique.
The co-occurrence matrix is created by performing PPR on the original graph and on the graph with all edges reversed, after which the co-occurrence columns, representing the counts for one entity, are nominalized.

From the previous step, we also obtain a list of all entities and predicates of the graph. 
This list is used for the Named Entity Recognition (NER) step, which in our current setup is very rudimentary (we only perform a simple string matching with the label of the entity, which will obviously lead to many errors and missed entities).
All English words, that did not match a resource identifier of an entity at the end of this step, are collected and attached to the entity-predicate word list. 
Now GloVe co-occurrence for text is directly applied to the modified text, and uses the combined entity-predicate and word list as an input.

Now, we have two co-occurrence matrices and a merging must be performed to be able to perform a joint embedding.
The KGloVe matrix is already normalized, but the other one is not.
So, we normalize by dividing all entries in the matrix by the 100th largest entry. 
After that the matrices are summed together.

\section{Experiment}
We evaluate the presented approach on classification and regression tasks, which were also used for the evaluation of KGloVe~\cite{KGlove}.
We use the entity embeddings corresponding to entities in five different datasets from different domains to predict classes or properties of \textit{Cities}, the \textit{Metacritic Movies}, the \textit{Metacritic Albums}, the \textit{AAUP} (using the average salary as target variable) and the \textit{Forbes} datasets.\footnote{\url{https://bit.ly/2J8WBdN}}
The results are calculated using stratified 10-fold cross validation.

%

\begin{table}[]
	\centering
	\caption{The average ranking of the classifier (resp. regression model) when evaluated 10 times over 5 different classification (resp. regression) tasks for different algorithms.}
	\label{resultstable}
	\begin{tabular}{@{}llllllllll@{}}
		& \multicolumn{5}{l}{Classification}        & \phantom{abc} &\multicolumn{3}{l}{ Regression }     \\ 
		& NB  & KNN  & SVM-100 & SVM-1000 & C45  &  & LR & KNN  & M5  \\ 
		\cmidrule{2-6} \cmidrule{8-10}
		KGloVe with literals & 1.6 & 1.74 & 1.2       & 1.38       & 1.76 &  & 1.6        & 1.46 & 1.6 \\ 
		Normal KGloVe   & 1.4 & 1.26 & 1.8       & 1.62       & 1.24 &  & 1.4        & 1.54 & 1.4 \\ 
		\bottomrule
	\end{tabular}
\end{table}

We can see from the table\footnote{Complete results from \url{https://bit.ly/2H9eS8J}} that there neither of the models consistently outperforms the other.
The KGloVe with literals does obtain the best rankings for the SVM models for classification, but for the regression results, the models perform very similarly.
The comparison is somewhat unfair, as we compare one of the best performing KGloVe models, with a first experiment for the approach including literals.
Likely, we can improve these results a lot by performing parameter tuning on a larger set of experiments and tasks.

\section{Conclusion and Outlook}

In this paper we presented a preliminary result from our ongoing investigation with a new model to incorporate the information contained in literals in a Knowledge Graph into the embedding. 
The experiment shows that there might be an advantage from incorporating literal information into latent feature models, but the current gain is rather small.
Many improvements of the presented technique are possible, like for example a better NER, a more balanced way to combine the graph and text co-occurrence matrices, inclusion of larger text corpora, and a large experimental analysis of the many parameters used in this model to determine good values.

\bibliographystyle{splncs04}
\bibliography{mybibliography}

\end{document}